\documentclass[twoside,11pt]{article}

\usepackage{jmlr2e_modified}
\usepackage{color}
\usepackage{enumitem}
\usepackage{url}

\usepackage{xspace} % Correct macro spacing
\def\mllib{\textsf{MLlib}\xspace}
\def\mlbase{\textsc{MLbase}\xspace}

\jmlrheading{}{}{}{}{}{Meng et al.}

\ShortHeadings{\mllib: Machine Learning in Apache Spark}{Meng et al.}
\firstpageno{1}

\begin{document}

\title{\mllib: Machine Learning in Apache Spark}

\author{\name Xiangrui Meng\textsuperscript{$\dagger$} \email meng@databricks.com \\
      \addr Databricks, 160 Spear Street, 13th Floor, San Francisco, CA 94105
      \AND
       \name Joseph Bradley \email joseph@databricks.com \\
      \addr Databricks, 160 Spear Street, 13th Floor, San Francisco, CA 94105
      \AND
      \name Burak Yavuz \email burak@databricks.com \\
      \addr Databricks, 160 Spear Street, 13th Floor, San Francisco, CA 94105
     \AND
       \name Evan Sparks \email sparks@cs.berkeley.edu \\
       \addr UC Berkeley, 465 Soda Hall, Berkeley, CA 94720
       \AND
       \name Shivaram Venkataraman \email shivaram@eecs.berkeley.edu \\
       \addr UC Berkeley, 465 Soda Hall, Berkeley, CA 94720
       \AND
      \name Davies Liu \email davies@databricks.com \\
      \addr Databricks, 160 Spear Street, 13th Floor, San Francisco, CA 94105
     \AND
       \name Jeremy Freeman \email freemanj11@janelia.hhmi.org \\
       \addr HHMI Janelia Research Campus, 19805 Helix Dr, Ashburn, VA 20147
       \AND
      \name DB Tsai \email dbt@netflix.com \\
      \addr Netflix, 970 University Ave, Los Gatos, CA 95032
     \AND
       \name Manish Amde \email manish@origamilogic.com \\
       \addr Origami Logic, 1134 Crane Street, Menlo Park, CA 94025 
      \AND
       \name Sean Owen \email sowen@cloudera.com \\
       \addr Cloudera UK, 33 Creechurch Lane, London EC3A 5EB United Kingdom
      \AND
       \name Doris Xin \email dorx0@illinois.edu \\
       \addr UIUC, 201 N Goodwin Ave, Urbana, IL 61801
      \AND
      \name Reynold Xin\email rxin@databricks.com \\
      \addr Databricks, 160 Spear Street, 13th Floor, San Francisco, CA 94105
     \AND
       \name Michael J. Franklin \email franklin@cs.berkeley.edu \\
       \addr UC Berkeley, 465 Soda Hall, Berkeley, CA 94720
       \AND
       \name Reza Zadeh \email rezab@stanford.edu \\
      \addr Stanford and Databricks, 475 Via Ortega, Stanford, CA 94305
      \AND
       \name Matei Zaharia \email matei@mit.edu \\
       \addr MIT and Databricks, 160 Spear Street, 13th Floor, San Francisco, CA 94105
       \AND
       \name Ameet Talwalkar\textsuperscript{$\dagger$} 
       \email ameet@cs.ucla.edu \\
       \addr UCLA and Databricks, 4732 Boelter Hall, Los Angeles, CA 90095
}

\editor{}

\maketitle
{
\vspace*{-0.8cm}
\noindent\textsuperscript{$\dagger$} Corresponding authors.\par\bigskip
\par
}
\newpage
\begin{abstract}%   <- trailing '%' for backward compatibility of .sty file
Apache Spark is a popular open-source platform for large-scale data processing
that is well-suited for iterative machine learning tasks.  In this paper we
present \mllib, Spark's open-source distributed machine learning library.
\mllib provides efficient functionality for a wide range of learning settings 
%including classification, regression, collaborative
%filtering, clustering, and dimensionality reduction, 
and includes several underlying statistical, optimization, and linear
algebra primitives. Shipped with Spark, % which provides APIs for Java, Scala,
%Python, and R, 
\mllib supports several languages and provides a high-level API that leverages
Spark's rich ecosystem to simplify the development of end-to-end machine
learning pipelines. \mllib has experienced a rapid growth due to its vibrant
open-source community of over 140 contributors, and includes extensive
documentation to support further growth and to let users quickly get up to speed.
\end{abstract}

\begin{keywords}
  Scalable Machine Learning, Distributed Algorithms, Apache Spark
\end{keywords}

\section{Introduction}
Modern datasets are rapidly growing in size and complexity, and there is a
pressing need to develop solutions to harness this wealth of data using
statistical methods.  Several `next generation' data flow engines that
generalize MapReduce~\citep{MapReduce} have been developed for large-scale data
processing, and building machine learning functionality on these engines is a
problem of great interest.  In particular, Apache Spark~\citep{Spark} has
emerged as a widely used open-source engine.  Spark is a fault-tolerant and
general-purpose cluster computing system providing APIs in Java, Scala, Python,
and R, along with an optimized engine that supports general execution graphs.
Moreover, Spark is efficient at iterative computations and is thus
well-suited for the development of large-scale machine learning applications.
 
In this work we present \mllib, Spark's distributed machine learning library, and
the largest such library.  The library
targets large-scale learning settings that benefit from data-parallelism or
model-parallelism to store and operate on data or models. \mllib 
consists of fast and scalable implementations of standard
learning algorithms for common learning settings including
classification, regression, collaborative filtering, clustering, and
dimensionality reduction. It also provides a variety of underlying
statistics, linear algebra, and optimization primitives.
%that support data analysis,
Written in Scala and using native (\verb!C++! based) linear algebra libraries on each
node, \mllib includes Java, Scala, and Python APIs, and is released
as part of the Spark project under the Apache 2.0 license.

\begin{figure}
\centering
\includegraphics[width=0.40\textwidth,scale=0.45]{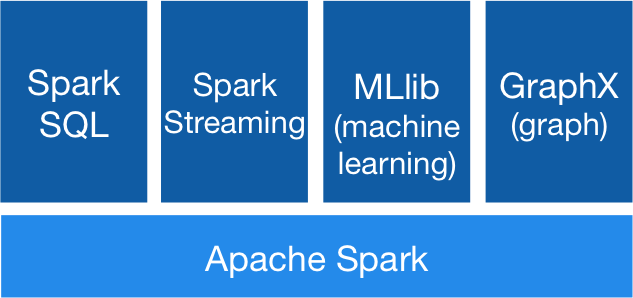}
\includegraphics[width=0.43\textwidth,scale=0.45]{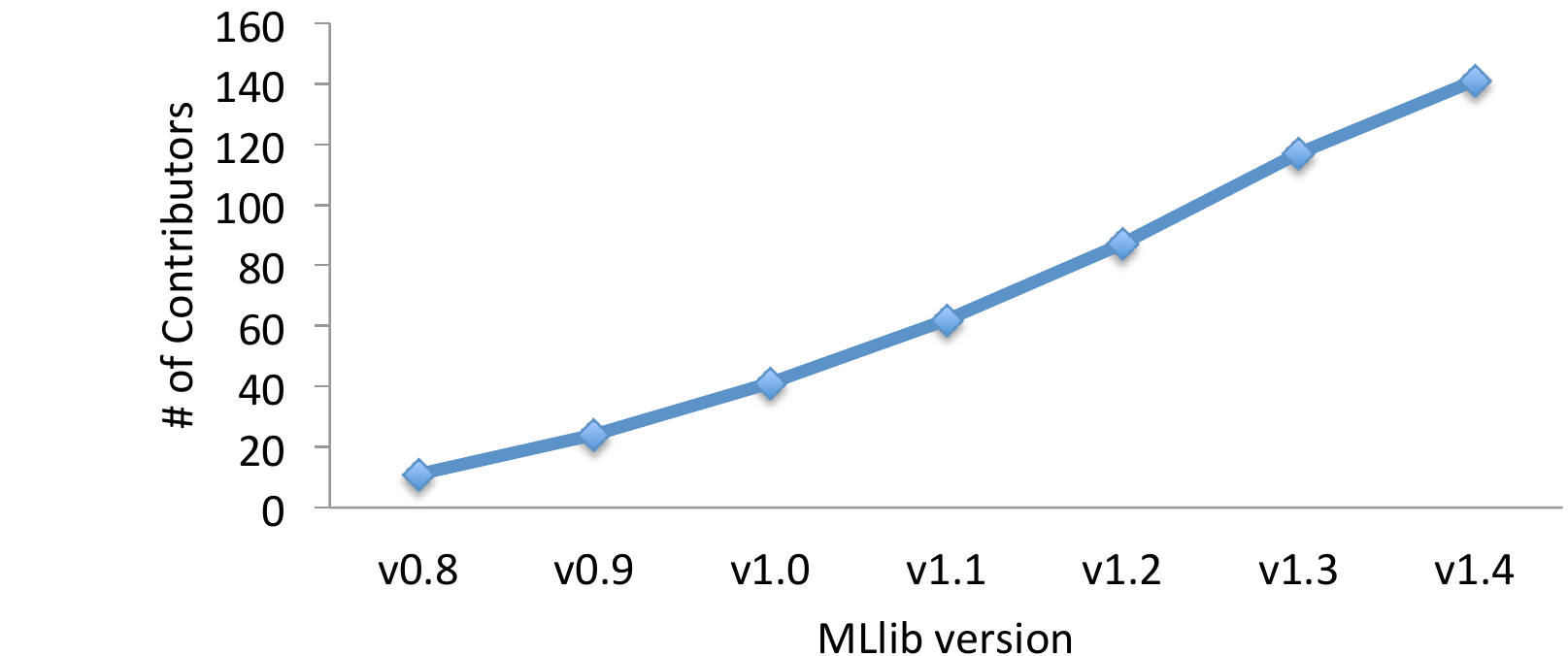} \\
 \qquad(a) \qquad\qquad\qquad\qquad\qquad\qquad\qquad\qquad\qquad(b)
\caption{(a) Apache Spark ecosytem. (b). Growth in \mllib contributors. }
%\caption{(a) Apache Spark Ecosytem.  Higher-level libraries, including \mllib,
%are tightly integrated with the underlying Spark engine and packaged together
%in code releases. (b).  \mllib Contributors. \mllib is under active development
%with a rapid growth in open-source contributors from a wide range of
%organizations since its original release with Spark version 0.8 in September
%2013. \todo{update plot with real numbers.} }
\label{fig:mllib_info}
\end{figure}

\mllib's tight integration with Spark results in several benefits.  First,
since Spark is designed with iterative computation in mind, it enables the
development of efficient implementations of large-scale machine learning
algorithms since they are typically iterative in nature. 
Improvements in low-level components of Spark often translate into performance
gains in \mllib, without any direct changes to the library itself. Second,
Spark's vibrant open-source community has led to rapid growth and adoption of
\mllib, including contributions from over 140 people.  Third, \mllib is one of
several high-level libraries built on top of Spark, as shown in
Figure~\ref{fig:mllib_info}(a). As part of Spark's rich ecosystem, and in part
due to \mllib's \texttt{spark.ml} API for pipeline development, \mllib provides
developers with a wide range of tools to simplify the development of machine
learning pipelines in practice.

%In the remainder of this paper we ...

%\begin{itemize}
%\item Intro
%\item Growing importance of Distributed Computing
%\item Spark becoming de facto platform for large-scale distributed computing
%\item Explain what Spark is \cite{Spark}
%\item MLlib leverages Spark environment
%\item Apache License
%\item Fastest growing open-source distributed ML library
%\item Supports wide range of ML functionality
%\end{itemize}

\section{History and Growth}
Spark was started in the UC Berkeley AMPLab and open-sourced in 2010.  
Spark is designed for efficient iterative computation and starting with
early releases has been 
packaged with example machine learning algorithms.  However, it lacked a suite of
robust and scalable learning algorithms until the creation of \mllib.
Development of \mllib began in 2012 as part of the \mlbase
project~\citep{MLbase}, and \mllib was open-sourced in September 2013.
From its inception, \mllib has been packaged with Spark, with
the initial release of \mllib included in the Spark 0.8 release. As an Apache
project, Spark (and consequently \mllib) is open-sourced under the Apache 2.0
license. Moreover, as of Spark version 1.0, Spark and \mllib are on a
3-month release cycle.

The original version of \mllib was developed at UC Berkeley by $11$ contributors,
and
%The initial release of \mllib 
provided a limited set of standard machine
learning methods.
%, including linear models for classification and regression
%trained via distributed stochastic gradient descent, along with scalable
%implementations for $k$-means clustering and collaborative filtering via
%alternating least squares. The library included unit tests and
%associated data generators to verify algorithm correctness, and initially supported
%Java and Scala APIs.
Since this original release, \mllib has experienced dramatic growth in terms of
contributors. 
Less than two years later, as of the Spark 1.4 release, \mllib has over 140 
contributors from over 50 organizations. Figure~\ref{fig:mllib_info}(b) demonstrates the growth in \mllib's open source
community as a function of release version.  The strength of this open-source
community has spurred the development of a wide range of additional functionality.
%For instance, \mllib version 0.9 added a Python API; version 1.0 included a
%variety of new learning algorithms and optimization primitives (e.g., decision
%trees, principal component analysis, L-BFGS) along with support for sparse
%data; and version 1.1 incorporated advanced distributed communication
%primitives as well as several new learning
%methods.
% (including streaming linear regression).  See
%Section~\ref{sec:functionality} for a more detailed description of \mllib
%functionality.
 
%\item part of MLbase project with 11 contributors from AMPLab \cite{MLbase}
%Java / Scala support
%list algorithms
%\item Describe evolution over time
%growing number of contributors from different organizations
%added Python support
%list various algorithms / functionality
%Can show via figures, e.g., of number of contributors over time, visualizations of MLlib history in Xiangrui’s NYC ML Meetup on 3/19/15
%\item Spark Packages
%For domain specific algorithms, another venue for contributing
%(note for self: spark packages used for 3 things: Data sources, ML, examples/tutorials)
%\end{itemize}

\section{Core Features}
\label{sec:functionality}
In this section we highlight the core features of \mllib; we refer the reader to
the \mllib user guide for additional details~\citep{mllib_guide}.

%\begin{itemize}[leftmargin=*]
\emph{Supported Methods and Utilities}. \mllib provides fast, distributed
implementations of common learning algorithms,
%including\footnote{This is not an exhaustive list.}: various linear models,
including (but not limited to): various linear models,
naive Bayes, and ensembles of decision trees for classification and regression
problems; alternating least squares with explicit and implicit feedback for
collaborative filtering; and $k$-means
clustering and principal component analysis for clustering and dimensionality
reduction. The library also provides a number of low-level primitives and basic
utilities for convex optimization, distributed linear algebra, statistical
analysis, and feature extraction, and supports various I/O formats, including
native support for LIBSVM format, data integration via
Spark SQL~\citep{SparkSQL}, as well as PMML~\citep{pmml} and \mllib's internal
format for model export. 

\emph{Algorithmic Optimizations}.
\mllib includes many optimizations to support efficient distributed learning
and prediction.  We highlight a few cases here.  The ALS algorithm for
recommendation makes careful use of blocking to reduce JVM garbage collection
overhead and to utilize higher-level linear algebra operations.  Decision trees
use many ideas from the PLANET project~\citep{Planet}, such as data-dependent
feature discretization to reduce communication costs, and tree ensembles
parallelize learning both within trees and across trees.  Generalized linear
models are learned via optimization algorithms which parallelize gradient
computation, using fast \verb!C++!-based linear algebra libraries for worker
computations.  Many algorithms benefit from efficient communication primitives;
in particular tree-structured aggregation prevents the driver from being a
bottleneck, and Spark broadcast quickly distributes large models to workers.

\emph{Pipeline API}.
Practical machine learning pipelines often involve a sequence of data
pre-processing, feature extraction, model fitting, and validation stages. 
%For %example, classifying text documents might involve text segmentation and
%cleaning, extracting features, and training a classification model with
%cross-validation.  
Most machine learning libraries do not provide native support for the diverse
set of functionality required for pipeline construction.  Especially when
dealing with large-scale datasets, the process of cobbling together an
end-to-end pipeline is both labor-intensive and expensive in terms of network
overhead. Leveraging Spark's rich ecosystem and inspired by previous
work~\citep{scikitlearn,SKlearnAPI,MLI,sparks2015tupaq}, \mllib includes a
package aimed to address these concerns. This package, called
\texttt{spark.ml}, simplifies the development and tuning of multi-stage
learning pipelines by providing a uniform set of high-level
APIs~\citep{pipelines}, including APIs that enable users to swap out a standard
learning approach in place of their own specialized algorithms.

\emph{Spark Integration}.
\mllib benefits from the various components within the Spark ecosystem.  At the
lowest level, Spark core provides a general execution engine with over 80
operators for transforming data, e.g., for data cleaning and featurization.
\mllib also leverages the other high-level libraries packaged with Spark.
Spark SQL provides data integration functionality, SQL and structured data
processing which can simplify data cleaning and preprocessing, and also
supports the DataFrame abstraction which is fundamental to the
\texttt{spark.ml} package.  GraphX~\citep{GraphX} supports large-scale graph
processing and provides a powerful API for implementing learning algorithms
that can naturally be viewed as large, sparse graph problems, e.g.,
LDA~\citep{LDA, bradleyLDA}.  Additionally, Spark Streaming~\citep{SparkStreaming}
allows users to process live data streams and thus enables the development of
online learning algorithms, as in \citet{freemanClustering}.  Moreover,
performance improvements in Spark core and these high-level libraries lead to
corresponding improvements in \mllib. 

\emph{Documentation, Community, and Dependencies}. The \mllib user guide
provides extensive documentation; it describes all supported methods and
utilities and includes several code examples along with API docs for all
supported languages~\citep{mllib_guide}.  The user guide also lists \mllib's
code dependencies, which as of version 1.4 are the following open-source
libraries: Breeze, netlib-java, and (in Python)
NumPy~\citep{breeze,netlib-java,jblas,numpy}.  Moreover, as part of the Spark
ecosystem, \mllib has active community mailing lists, frequent meetup events,
and JIRA issue tracking to facilitate open-source
contributions~\citep{spark-community}. To further encourage community
contributions, Spark Packages~\citep{spark-packages} provides a community
package index to track the growing number of open source packages and libraries
that work with Spark.  To date, several of the contributed packages consist of
machine learning functionality that builds on \mllib. Finally, a massive open
online course has been created to describe the core algorithmic concepts used
to develop the distributed implementations within \mllib~\citep{mooc}.  

\begin{figure}
\centering
\includegraphics[width=0.45\textwidth,scale=0.5]{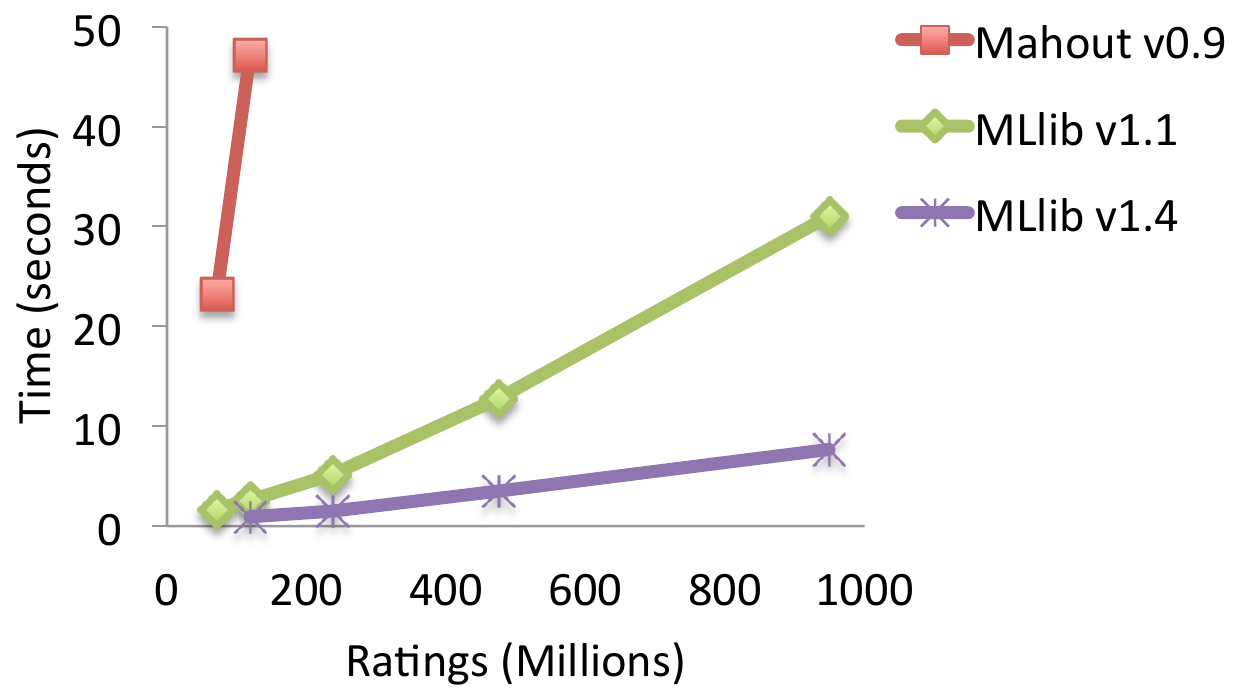}
\includegraphics[width=0.45\textwidth,scale=0.5]{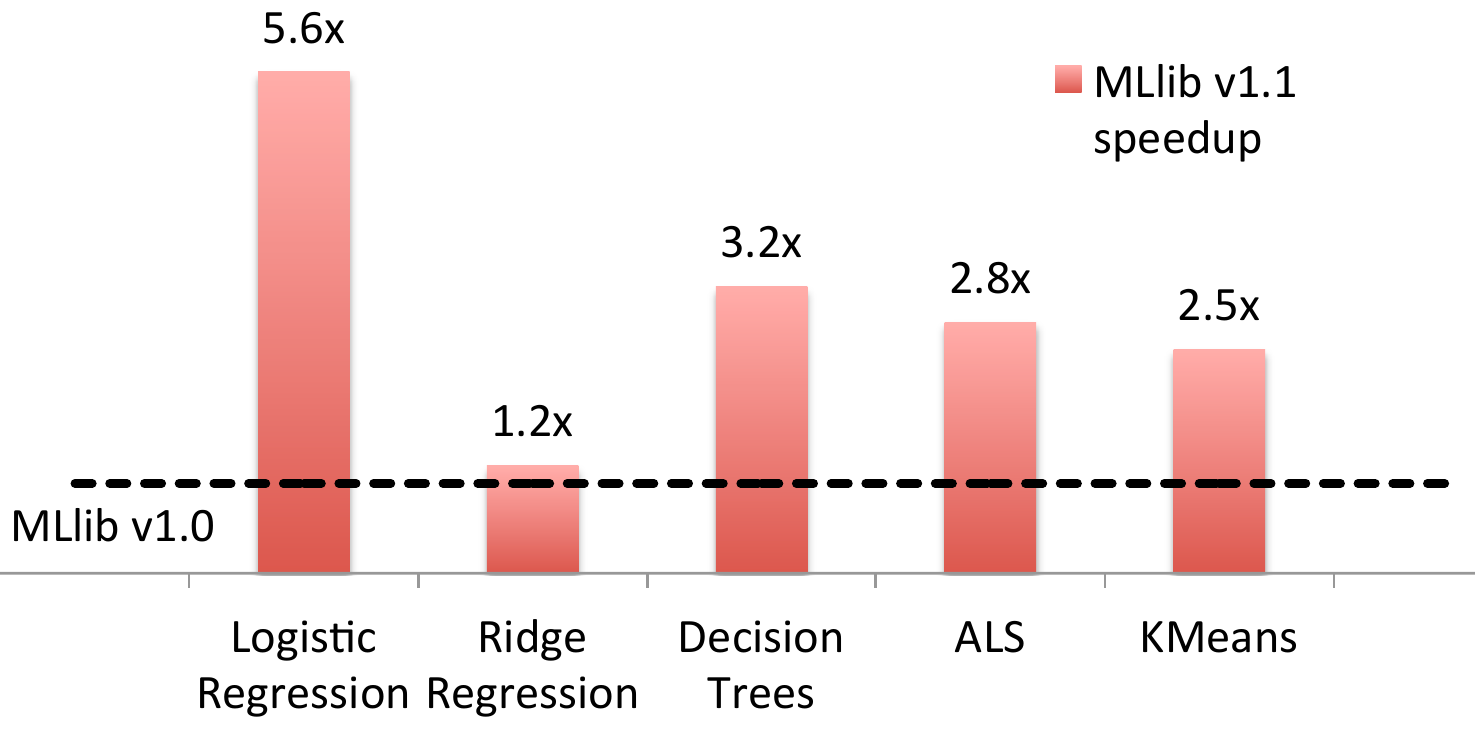} \\
(a) \qquad\qquad\qquad\qquad\qquad\qquad\qquad\qquad\qquad(b)
\caption{(a) Benchmarking results for ALS. (b) \mllib speedup between versions.}
\label{fig:performance}
\end{figure}

\section{Performance and Scalability}
In this section we briefly demonstrate the speed, scalability, and continued
improvements in \mllib over time.  We first look at scalability
by considering ALS, a commonly used collaborative filtering approach. 
For this benchmark, we worked with
scaled copies of the Amazon Reviews dataset~\citep{McAuley:2013}, where we
duplicated user information as necessary to increase the size of the data. We
ran $5$ iterations of \mllib's ALS for various scaled copies of the dataset,
running on a 16 node EC2 cluster with m3.2xlarge instances using \mllib
versions 1.1 and 1.4.  For comparison purposes, we ran the same experiment using Apache
Mahout version 0.9~\citep{mahout},
which runs on Hadoop MapReduce.  Benchmarking results, presented in
Figure~\ref{fig:performance}(a), demonstrate that MapReduce's scheduling overhead
and lack of support for iterative computation substantially slow down its
performance on moderately sized datasets.  In contrast, \mllib exhibits
excellent performance and scalability, and in fact can scale to much larger
problems.% with reported results on as many as 50 billion ratings.
%  Using 50 nodes, we ran 10 iterations of \mllib's ALS on a scaled
%version of the Amazon dataset containing 3.5 billion ratings, 660 million users, and
%2.4 million items in 40 minutes (with a cost of less than \$2 using EC2 spot
%instances). 

Next, we compare \mllib versions 1.0 and 1.1 to evaluate improvement over time.
We measure the performance of common machine learning methods in \mllib, with
all experiments performed on EC2 using m3.2xlarge instances with 16 worker
nodes and synthetic datasets from the spark-perf package
(\url{https://github.com/databricks/spark-perf}).  The results are presented in
Figure~\ref{fig:performance}(b), and show a 3$\times$ speedup on average across
all algorithms.  These results are due to specific algorithmic improvements (as
in the case of ALS and decision trees) as well as to general improvements to
communication protocols in Spark and \mllib in version 1.1~\citep{YavuzMeng}.

%\begin{itemize}
%\item Show ALS results comparing MLlib and Mahout
%\item Show plot between 1.0 and 1.1 showing benefits from integration with Spark
%\item highlighting that we get speedups for free as Spark Core improves
%\end{itemize}

% Acknowledgements should go at the end, before appendices and references
\section{Conclusion}
\mllib is in active development, and the following link
provides details on how to contribute:
\url{https://cwiki.apache.org/confluence/display/SPARK/Contributing+to+Spark}.
Moreover, 
we would like to acknowledge all \mllib contributors. 
The list of
Spark contributors can be found at \url{https://github.com/apache/spark}, and
the \texttt{git log} command can be used to identify \mllib
contributors.

\newpage
\bibliography{refs}

\end{document}